\documentclass{article}
\usepackage{arxiv}

\usepackage[utf8]{inputenc} 
\usepackage[T1]{fontenc}    
\usepackage{hyperref}       
\usepackage{url}            
\usepackage{booktabs}       
\usepackage{amsfonts}       
\usepackage{nicefrac}       
\usepackage{microtype}      
\usepackage{lipsum}		
\usepackage{graphicx}
\usepackage{natbib}
\usepackage{doi}
\usepackage{multirow}
\usepackage{multicol}

\usepackage[usenames,dvipsnames]{xcolor}

\title{The Claire French Dialogue Dataset
}
\author{Julie Hunter\thanks{Equal contributions.}\\
	LINAGORA\\
	Toulouse, France \\
	\texttt{jhunter@linagora.com} \\
	\And
    Jérôme Louradour$^\ast$ \\
	LINAGORA\\
	Toulouse, France \\
	\texttt{jlouradour@linagora.com} \\
    	\And
	Virgile Rennard \\
	LINAGORA\\
        Ecole Polytechnique\\
	Paris, France \\
	\texttt{vrennard@linagora.com} \\
    	\And
	Ismaïl Harrando \\
	LINAGORA\\
	Toulouse, France \\
	\texttt{iharrando@linagora.com} \\
    	\And
    Guokan Shang \\
	LINAGORA\\
	Paris, France \\
	\texttt{gshang@linagora.com} \\
 	\And
    Jean-Pierre Lorré \\
	LINAGORA\\
	Toulouse, France \\
	\texttt{jplorre@linagora.com} \\
}
\date{November 2023}

\renewcommand{\headeright}{}
\renewcommand{\undertitle}{}

\newcommand{\openlicense}{\href{https://www.etalab.gouv.fr/wp-content/uploads/2017/04/ETALAB-Licence-Ouverte-v2.0.pdf}{Open License 2.0}}
\newcommand{\ccbysa}{\href{https://creativecommons.org/licenses/by-sa/4.0/}{CC BY-SA}}
\newcommand{\ccbysatrois}{\href{https://creativecommons.org/licenses/by-sa/3.0/}{CC BY-SA}}
\newcommand{\usedapache}{$^\dagger$}

\begin{document}

\maketitle

\begin{abstract}
	We present the Claire French Dialogue Dataset (CFDD),\footnote{\href{https://huggingface.co/datasets/OpenLLM-France/Claire-Dialogue-French-0.1}{https://huggingface.co/datasets/OpenLLM-France/Claire-Dialogue-French-0.1}} a resource created by members of LINAGORA Labs\footnote{\href{https://labs.linagora.com/}{https://labs.linagora.com/}} in the context of the OpenLLM France initiative.\footnote{\href{https://huggingface.co/OpenLLM-France}{https://huggingface.co/OpenLLM-France}} CFDD is a corpus containing roughly 160 million words from  transcripts and stage plays in French that we have assembled and publicly released in an effort to further the development of multilingual, open source language models. This paper describes the 24 individual corpora of which CFDD is composed and provides links and citations to their original sources. It also provides our proposed breakdown of the full CFDD dataset into eight categories of subcorpora and describes the process we followed to standardize the format of the final dataset.    We conclude with a discussion of similar work and future directions. 
\end{abstract}


\section{Introduction}

The overwhelming success of OpenAI's ChatGPT, whose first version was released one year ago, has led to an undeniable surge of excitement about large language models (LLMs) among researchers and the general public alike. OpenAI's anything-but-open approach to sharing its models or information about training them, however, has led to an equally passionate reaction among those who feel that AI development should be widely accessible and that data usage should be transparent in order to protect the rights of those who have contributed the data and that data -- a resource crucial to the development and understanding of AI models -- should be shared with the broader research community.

The call for transparency has begun to bear fruit. High-profile language models like Falcon,\citep{falcon40b} LLaMa2 \citep{llama2} and MPT \citep{MosaicML} -- to name just a few -- come very close to a classic definition of open source. A central part of OpenLLM France's mission is to contribute to this momentum by building language models and remaining fully transparent about every step of model training, including the data used for training. Another objective, which we find equally important, is to increase the availability of language models and training data geared to languages other than English and to non-anglophone cultures. Indeed, the majority of the high-profile LLMs available today are trained primarily on English documents coming from anglophone cultures. Only 0.16\% of the data used to train LLaMa2 comes from French, for example. 

This paper presents our first step to addressing these objectives: the release of Claire French Dialogue Dataset (CFDD). 
CFDD is designed for NLP tasks requiring understanding and generation of spontaneous, oral French dialogue. This might include the development of chatbots and vocal assistants that can speak naturally and perhaps even adjust their conversational style to their audience, adopting a more informal conversational tone for someone calling a support line, or a more formal tone in a business context. Or it could involve training models for meeting summarization \citep{rennard2023abstractive} or question-answering based on transcripts.  Recent advances in speech processing (such as OpenAI's Whisper) make such exploitation of transcripts by NLP systems more accessible at a large scale than ever before.

From a linguistic point of view, however, the nature of spontaneous dialogue is quite unlike that of written, document-style texts that are often used for training language models.  A central part of dialogue is conversational exchange in interaction -- the way in which one utterance might be a reaction not only to something previously said but also to the person who said it, leading to a higher use of first and second person pronouns, feedback utterances (\textit{hmm-hmm}, \textit{ok}), question/answer sequences, disagreements and so on. The type of language involved is also often far more disfluent, with frequent repetitions, truncated words (\textit{to- today}) and filled pauses (\textit{um}). Our hypothesis is that by training on naturally occurring dialogue data to help a model build representations of the types of interactions common in conversations, we can build a solid foundation for downstream tasks that require understanding of this data.


\section{Related open data initiatives}
Numerous efforts have been made to assemble collections of French corpora that are distributed with open access for the sake of research.  CFDD has benefited from some of these, notably the platform \textit{Ortolang}\footnote{\href{https://www.ortolang.fr/fr/accueil/}{https://www.ortolang.fr/fr/accueil/}} (Plate-forme d'outils et de ressources linguistiques pour un traitement optimisé de la langue française -- Platform of linguistic tools and resources for  optimized processing of the French language), which hosts at least a version of the majority of the original corpora that make up CFDD. The Orféo platform (Outils et Ressources sur le Français Ecrit et Oral - Tools and Resources on Written and Spoken French)\footnote{Orféo platform: \href{http://ortolang107.inist.fr/}{http://ortolang107.inist.fr/}} hosts a variety of French corpora and tools for exploiting them, with an eye to supporting linguistic research. Projects such as the CEFC\footnote{\href{https://repository.ortolang.fr/api/content/cefc-orfeo/11/documentation/site-orfeo/home/index.html}{https://repository.ortolang.fr/api/content/cefc-orfeo/11/documentation/site-orfeo/home/index.html}} (Corpus d’Etude pour le Français Contemporain -
Corpus for the Study of Contemporary French) and Parole Publique \citep{otg_ubs} offer collections of text and oral corpora, sometimes with standardized formats.

Recent years have also witnessed the release of massive data collections designed specifically for the development of LLMs, whose ability to understand, process, and generate human-like text is directly proportional to the quality and quantity of the data on which they are trained.  The ROOTS (Responsible Open-science Open-collaboration Text
Sources) corpus \citep{roots}, assembled by the BigScience initiative and used to train the BLOOM language model \citep{bloom_short}, is a 1.6TB dataset including data from 498 corpora that span 59 languages, including French.\footnote{\href{https://huggingface.co/bigscience-data}{https://huggingface.co/bigscience-data}} RefinedWeb \citep{penedo2023refinedweb}, OSCAR \citep{suarez2019asynchronous}, and RedPajama \citep{together2023redpajama} are all created by filtering and refining large dumps of CommonCrawl data\footnote{\href{https://commoncrawl.org/}{https://commoncrawl.org/}} differently and preparing the data for LLM training. While these datasets contain data on French, with a special emphasis on multilinguality in the ROOTS corpus, they do not share the focus on dialogue that characterizes CFDD. 

In terms of dialogue-focused efforts, DialogStudio \citep{zhang2023dialogstudio} is a unified collection for conversational AI encompassing primarily English human-annotated dialogue datasets. Like many of the original datasets in CFDD, most of these corpora are annotated with extra information meant to help with downstream conversational tasks. 
The collection consists of 87 datasets, which are categorized into six groups: open-domain dialogues, task-oriented dialogues, natural language understanding tasks, conversational recommendation, dialogue summarization and knowledge-grounding dialogue. 
Other notable collections include InstructDial  \citep{gupta-etal-2022-instructdial} and ParlAI \citep{miller-etal-2017-parlai}, which have the same nature as DialogStudio, but on a smaller scale.

\section{The component datasets}\label{sec:datasets}
In an effort to make our training process as open and transparent as possible, we have decided to release the entirety of CFDD in the form that we used for training Claire-7B-0.1\footnote{\href{https://huggingface.co/OpenLLM-France/Claire-7B-0.1}{https://huggingface.co/OpenLLM-France/Claire-7B-0.1}} and Claire-Mistral-7B-0.1\footnote{\href{https://huggingface.co/OpenLLM-France/Claire-Mistral-7B-0.1}{https://huggingface.co/OpenLLM-France/Claire-Mistral-7B-0.1}} (see Section \ref{sec:data-preparation} for a description of our data preparation and normalization process). A number of the original corpora used were created for research purposes, and often meticulously prepared by linguists and annotators for specific scientific objectives. Most of these datasets are, understandably, released with a type of license restricted to research purposes, namely \href{https://creativecommons.org/licenses/by-nc-sa/4.0/deed.en}{CC BY-NC-SA}. Accordingly, the entirety of CFDD is also released under a CC BY-NC-SA license, though we note that a subset of the CFDD is available under more permissive licences and some of these were used to train the Claire-7B-Apache-0.1 model,\footnote{\href{https://huggingface.co/OpenLLM-France/Claire-7B-Apache-0.1}{https://huggingface.co/OpenLLM-France/Claire-7B-Apache-0.1}}. These corpora are indicated with {\usedapache} in Table \ref{tab:categories}. 

The data in CFDD come entirely from either stage plays, both classic plays and more modern works, or from naturally occurring French dialogues that have been transcribed. Section \ref{sec:data-description} describes in more detail the characteristics of the data in each corpus that we used. But first, we introduce the individual corpora here and provide a quick description of its context and objectives. 


\begin{itemize}

    \item[] {\bf ACSYNT} \citep{acsynt}: The ACSYNT corpus is an oral corpus of contemporary French recorded with speakers from southwest France. The original corpus is composed of three types of oral data: oral reading of text, monologue presentations, and guided interviews. CFDD includes only the subset of dialogues. ACSYNT was designed for study of prosody and intonation, oral syntax and discourse analysis.
    \item[] {\bf Assemblée Nationale} \citep{assemblee_nationale}: The website for the Assemblée Nationale (French National Assembly/Parliament) offers years of transcripts for a variety of parliamentary debates and discussions. CFDD includes a selection of transcripts from Sessions 13-16 that were scraped from the web. Transcripts were chosen based on criteria designed to identify more dialogical interaction. 
    \item[] {\bf Orféo-CEFC} \citep{orfeo1,orfeo2,orfeo3}: The CEFC includes both written and oral corpora. The oral data, some of which was used for CFDD, include interactions with adult speakers from around France, Switzerland and Belgium in a variety of contexts. Transcriptions have been carefully checked.
    \begin{itemize}
        \item {\bf C-ORAL-ROM} \citep{orfeo_coralrom}: The C-ORAL-ROM corpus is a multilingual resource that provides data from spontaneous speech from a variety of Romance languages, mainly French, Italian, Portuguese and Spanish. The resource is the result of the C-ORAL-ROM project, which was undertaken by a European consortium, coordinated by the University of Florence and funded under the EU's Fifth Framework Program. The entire French section was made available to the Orféo project and was used for CFDD.
        \item {\bf CRFP} \citep{orfeo_crfp}: The CRFP (Corpus de référence du français parlé - Reference Corpus for Spoken French) consists of 134 recordings collected in around forty different cities in France and sampled according to 3 speech situations and certain speaker characteristics (level of education, age, gender). The entire corpus is in transcribed form and has been made available to the Orféo project.
        \item {\bf FLEURON}: FLEURON includes multimedia resources representative of  situations that international students tend to encounter on their arrival at a French university: interactions between students and administrative staff (courses and programs, the university restaurant, etc.), between students and library staff, between students and teachers (interviews, oral exams, viva voce), interactions from daily life outside the university campus (at the bank, train station, etc.) and testimonials from international students. A subset of these interactions was made available to the Orféo project. 
        \item {\bf Valibel} \citep{orfeo_valibel}: The Valibel Discours et Variation center manages a textual database that includes multiple corpora from recordings of oral interactions with participants from Brussels and Wallonia. A small subset of these corpora was made available to the Orféo project and, thus, CFDD.
    \end{itemize}
    \item[] {\bf Orféo} (other)
    \begin{itemize}
        \item {\bf CFPB} \citep{cfpb}: The objective of CFPB (Corpus de Français Parlé à Bruxelles - French as Spoken in Brussels) is parallel to that of CFPP, namely to create a corpus of interviews that illustrate the way that French is spoken in informal situations in Brussels at the beginning of the 21st century. 
        \item {\bf Réunions De Travail} (Work Meetings): We have been unable to find a document describing this corpus, but the meetings appear to be real, professional meetings.
    \end{itemize}
    \item[] {\bf CFPP} \citep{cfpp1,cfpp2}: CFPP (Corpus de Français Parlé Parisien des Années 2000 - Corpus of Spoken, Parisian French in the 2000s) is a collection of interviews recorded on the street  in Paris. The objective was to represent French the way it was spoken in informal situations in the beginning of the 21st century. This corpus is also distributed through the Orféo platform. 
    \item[] {\bf CID} \citep{cid1,cid2,cid3}: CID (Corpus of Interactional Data) is a corpus of one-hour long conversations between two people on open topics. The corpus offers a variety of annotations including phonetic transcription, disfluencies and discourse units.
    \item[] {\bf CLAPI} \citep{clapi}: CLAPI (Corpus de LAngue Parlée en Interaction - Corpus of Spoken Language in contexts of Interaction) is a multimedia database of informal conversations recorded in a variety of real-life situations, including: professional, institutional or private, commercial, didactic and medical interactions. A subset of this corpus is also distributed through the Orféo platform. 
    \item[] {\bf ESLO} \citep{elso}: ESLO (Enquête Sociolinguistique à Orléans - Sociolinguistic Study in Orléans) was a project led by the Laboratoire Ligérien de Linguistique (UMR7270) at the University of Orléans. ESLO 1 had a didactic purpose: to produce interviews and conversations, both in real life and on the telephone, that represented common social situations to help foreign speakers of French learn the language. ESLO 2, an effort started 40 years after the beginning of ESLO 1, builds on ESLO 1. CFDD includes all categories of the ESLO 1 and ESLO 2 datasets except \texttt{ESLO\_LIVRENF} (``livres enfants'' - children's books), which contains transcripts of children reading books with the help of adults. 
    \item[] {\bf FREDSum} \citep{fredsum}: FREDSum (French Debates) includes transcriptions from televised French political debates. The full corpus includes a variety of summmaries for each topic of each debate. 
    \item[] {\bf LinTO} \citep{linto}: LinTO includes transcripts from a variety of different types of interactions: meetings of a speech-processing team, presentations on various AI topics followed by Q/A sessions, short informal conversations, a role playing game.
    \item[] {\bf OFROM} \citep{ofrom}: OFROM (Corpus Oral de Français de Suisse ROMande - Oral Corpus of Swiss-Romande French) includes  excerpts from guided interviews of Swiss-born speakers living in the French-speaking part of Switzerland and more informal interactions. Topics range from jobs and travel to locals' hobbies, as well as their relationships with neighbors,  projects or incongruous situations they have encountered in their lives. They may also relate to Switzerland's political system or linguistic situation, etc. CFDD includes only transcripts that include both speakers in a dialogue. We found that many of the interviews excluded the transcript for the interviewer, which often led to very short monologues without context, and so we decided to exclude such cases. 
    \item[] {\bf Parole Publique}
    \begin{itemize}
        \item {\bf Accueil UBS} \citep{otg_ubs,ubs2}: The Accueil UBS (UBS reception desk) corpus is a pilot corpus of short calls between anonymous callers and the telephone reception staff in the Science and Law departments of the University of Southern Brittany (Université de Bretagne Sud). It was recorded by VALORIA in real-life conditions, following a semi-clandestine procedure in which only the reception staff were aware of the experiment, but were not given instructions on how to proceed. 
        \item {\bf OTG} \citep{otg_ubs,otg2}: The OTG corpus (Office du Tourisme de Grenoble - Grenoble Tourist Office)  is a pilot corpus of short, in-person dialogues between a tourist and a staff member of the Grenoble Tourist Office. The corpus was recorded by CLIPS-IMAG under real-life conditions, following a semi-automatic procedure in which only the office staff were aware of the recording until after the interaction.  No special instructions were given to reception staff.
    \end{itemize}

    \item[] {\bf Paris Stories} \citep{paris_stories}: Paris Stories is a corpus of oral French that contains monologues and dialogues from speakers living in the Parisian region. Students were given the assignment of interviewing a friend or relative and asking them to share a story about a given theme. The corpus provides a resource for studying contemporary spoken French and for training a syntactic parser. As such, the data has been morpho-syntactically annotated. 
    \item[] {\bf PFC} \citep{pfc1,pfc2}: PFC (Phonologie du Français Contemporain - Phonology of Contemporary French) contains a series of dialogues in which an interviewer engages the same subject in both a guided conversation and a more free-form conversation. The goal of the corpus is to serve as a resource for students learning French as a second language, but also for researchers in a variety of fields of linguistics including phonetics, phonology, and syntax as well as more conversational and pragmatic aspects of language. 

    \item[] {\bf Rhapsodie} \citep{rhapsodie,cfpp1,rhapsodie_avanzi,rhapsodie_lacheret,rhapsodie_mertens,rhapsodie_cprom,elso,pfc1}: Rhapsodie consists of short samples, some coming from external corpora, some coming from resources specific to Rhapsodie. The samples were selected to cover a variety of communicative contexts and genres, including monologue and dialogue, interviews and talk shows, and differing levels of interactive speech. The corpus contains phonetic, syntactic and prosodic annotations.
    \item[] {\bf SUMM-RE} \citep{summre}: The SUMM-RE corpus is composed of a series of 20 minute conversations designed to imitate certain features of real-life meetings: round-table reporting, decision making and planning. While participants were given guidance on how to structure the conversations, there was a lot of freedom in the final structure and topic of discussion, leading to a high level of spontaneous interaction. This corpus has not been manually transcribed, CFDD includes transcripts produced with a Whisper-based ASR pipeline\,\citep{whisper,whisperT} designed to handle spontaneous speech with multiple audio channels.
    \item[] {\bf TCOF} \citep{tcof}: The TCOF corpus (Traitement de Corpus Oraux en Français - Processing of Oral French Corpora) is the result of an effort to preserve older corpora for research by transcribing them. It contains both interactions between adults and children and interactions between adults. CFDD contains only the latter, as many of the interactions with children seemed to us hard to understand or to involve interactions other than pure dialogue, such  as an adult helping a child through a task. A version of the TCOF corpus is also distributed through the Orféo platform.
    \item[] {\bf Théâtre Classique} \citep{theatre_classique1,fredracor,dracor}: A collection of classic French plays dating back to the 15th century.
    \item[] {\bf Théâtre Gratuit} \citep{theatre_gratuit}: A collection of more modern French plays.
    
\end{itemize}

\textbf{A note on monologues}: While we made the decision to exclude monologues from certain datasets, such as ACSYNT and OFROM, as explained above, we opted to keep them for others. Rhapsodie, Paris Stories, Orféo/C-ORAL-ROM, Orféo/CRFP and TCOF (adults) contain a number of transcripts in which a speaker is responding to a single question that was clearly asked at the beginning of the recording but not transcribed, yielding monologues.  We  decided to keep these because we judged the transcripts to be sufficiently long, in contrast to OFROM, and the style of language involved was still very conversational. For the same reasons, we also decided to keep certain formal addresses, in which one speaker addresses an audience that does not speak back. These come notably from ESLO and Orféo/Valibel.

\section{Categories and general characteristics}\label{sec:data-description}


Our datasets come from a variety of different types of interactions, ranging from free conversations to guided interviews to more structured political debates. These corpora thus vary among numerous dimensions. Free conversations tend to have more informal language, for example, while political debates are more formal. Interviews tend to have a clearer question/response structure, while free conversations can meander. Because downstream tasks might benefit from being able to focus on certain types of interactions  more than others or, conversely, by having a roughly even representation of different types of interactions, we decided to group the corpora in CFDD by type of interaction. 

Grouping the corpora by interaction type led us to divide some of the original corpora into subcorpora. This task was greatly facilitated by corpora that used different  file naming conventions to signal different types of interactions. In the PFC corpus, for example, file names for the transcripts of guided interviews end with a ``g'' (before the suffix), while free (``libre'' in French) conversations end with an ``l''. For corpora in which different kinds of interactions were not easily retrievable from file names, we manually checked a random distribution of samples. This means that the categories are certainly far from perfect and future work could lead to further refining. 

The breakdown of the different corpora into categories, as well as certain statistics on the different subcorpora, are provided in Table \ref{tab:categories}. Further information on exactly which files were placed in each category can be found on Hugging Face.\footnote{\href{https://huggingface.co/datasets/OpenLLM-France/Claire-Dialogue-French-0.1/blob/main/FR/metadata_filter_datasets_regex.json}{\scriptsize https://huggingface.co/datasets/OpenLLM-France/Claire-Dialogue-French-0.1/blob/main/FR/metadata{\textunderscore}filter{\textunderscore}datasets{\textunderscore}regex.json}} Some of the categories are mostly straightforward. Parliamentary proceedings are multi-party interactions in which members of parliament discuss government issues. The theater pieces are quite diverse in terms of topic, but all share a theatrical style. Debates are yet another type of multi-party interaction, but tend to be relatively structured in that a moderator controls how participants take turns presenting their points of view on a given issue. Meetings are multi-party dialogues focused usually on sharing information or making decisions. Assistance dialogues tend to be between two people, and tend to involve short, problem-solving interactions.

The distinction between interviews, free conversation and presentations/formal addresses is relatively less clear. Presentations are, as the name suggests, presentations by a single speaker to an audience, but often the presentations involve extended question/answer interactions with the audience, making them resemble conversations. By contrast, formal addresses, such as a university president making a speech, involve little to no interaction with the audience. In the interview category, we privileged interactions with clear question/answer sequences in which one person was clearly an interviewer guiding the conversation. This means that despite the fact that Paris Stories, for example, is described as a corpus of interviews, we ended up placing it among the free conversations. As explained in Section \ref{sec:datasets}, many of the transcripts that we recovered from Paris Stories contain only a single long answer to a question that does not itself appear in the transcript. We decided to count these as free conversations rather than presentations because they were not designed as presentations.

\begin{table}
    \centering
    \begin{tabular}{lrrrl}
        \bf Corpus & \bf Words & \bf Turns & \bf Convs. & \bf License
        \\
        \hline
        \multicolumn{5}{c}{\textit{Parliamentary proceedings}} \\
        \hline
        Assemblée Nationale & 133M & 1.6M & 4. 5k & \openlicense\usedapache\\
        \hline 
        \multicolumn{5}{c}{\textit{Theater}}\\
        \hline
        Théâtre Classique & 12.8M & 441k & 25k\\
        Théâtre Gratuit & 2.7M & 155k & 4k & {\it no license}\usedapache\\
        \hline 
        \multicolumn{5}{c}{\textit{Interviews}}\\
        \hline
        ESLO (1/5) & 4.2M & 329k & 399\\
        TCOF (adults) & 765k & 49k & 237\\
        CFPP & 608k & 48k & 42\\
        ORFEO/Valibel (1/2) & 458k & 19k & 67\\
        PFC (1/2) & 268k & 15k & 173\\
        ORFEO/CFPB & 138k & 11k & 12\\
        ACSYNT & 61k  & 2.7k  & 144 & \ccbysa\\
        \hline 
        \multicolumn{5}{c}{\textit{Free conversations}}\\
        \hline
        OFROM & 590k & 44k & 151 \\
        ESLO (2/5) & 480k & 47k & 98\\
        ORFEO/CRFP & 405k & 9k & 124 \\ 
        ORFEO/C-ORAL-ROM & 248k & 6k & 152\\
        PFC (2/2) & 230k & 14k & 146\\
        CLAPI  & 122k & 15k & 14\\
        CID & 118k & 9k & 8\\
        Rhapsodie   & 28k & 1k & 41\\
        Paris Stories & 28k & 351 & 54 & \ccbysa\\
        LinTO (1/3) & 26k  & 2k & 4 & \ccbysa\usedapache\\
        \hline 
        \multicolumn{5}{c}{\textit{Meetings}}\\
        \hline
        SUMM-RE & 1.3M & 39k & 283 & \ccbysa\usedapache\\
        ORFEO/Reunions-de-Travail & 210k & 12k & 29\\
        LinTO (2/3) & 41k & 1.8k & 6 & \ccbysa\usedapache\\
        \hline 
        \multicolumn{5}{c}{\textit{Debates}}\\
        \hline
        FREDSum & 406k & 7k	& 144 & \ccbysa\usedapache\\
        ESLO (3/5) & 76k & 2k & 4 \\
        \hline 
        \multicolumn{5}{c}{\textit{Assistance}}\\
        \hline
        ESLO (4/5) & 95k &	11k &	143\\
        ORFEO/Fleuron & 33k & 2k & 51 \\
        OTG & 27k & 4k & 315 & \ccbysatrois \\
        Accueil UBS & 7.2k	& 1k & 41 & \ccbysatrois \\
        \hline 
        \multicolumn{5}{c}{\textit{Presentation, Formal address}}\\
        \hline
        ESLO (5/5) & 43k &	120 & 9 \\
        LinTO (3/3) & 38k	& 1.5k & 4 & \ccbysa\usedapache\\
        ORFEO/Valibel (2/2) & 12k &	5 & 5 \\
        & & & \\

    \end{tabular}
    \caption{
        Breakdown of subcorpora by category,
        with amounts of data (numbers of words, speech turns, and conversations), and licenses.
        When not indicated, licenses are
        \href{https://creativecommons.org/licenses/by-nc-sa/4.0/}{CC BY-NC-SA}. When the license is followed by \usedapache, it means that we used the corpus to train the \href{https://huggingface.co/OpenLLM-France/Claire-7B-Apache-0.1}{Claire-7B-Apache-0.1} model.
    }
    \label{tab:categories}
\end{table}

In addition to the statistics noted in Table \ref{tab:categories}, for the corpora for which we have access to the audio files, we determined that overall, the number of words per hour seems to be in the range of 10-15 thousand. 

\section{Data preparation}\label{sec:data-preparation}
Just as the corpora in CFDD come from a variety of different sources and were designed for diverse purposes, the formats in which they have been packaged and distributed vary quite wildly as well. This section describes the approach adopted to normalize the data format, starting with a presentation of some of the more significant differences between the datasets. We conclude with a discussion of the splits proposed for training and testing. 

\subsection{Diversity of the original data sets}

The original corpora have been distributed in differing formats, including \texttt{xml}, \texttt{json}, \texttt{TexGrid}, \texttt{txt}, \texttt{pdf} and \texttt{html}. Documents can  also vary significantly with regard to their internal structure. The \texttt{html} documents that had to be scraped from the Assemblée Nationale site, for instance, had such different formats from one legislative period to the next that we had to implement a number of ad-hoc post-processing heuristics, including some rules specific to individual legislative periods, to ensure a robust and consistent translation from the scraped \texttt{html} to raw text.

Speaker-identifying information and speaker turns are also  heterogeneous. Speaker labels are provided in different formats and are sometimes stored in files separate from the transcripts. In the latter case, files containing speaker labels had to be merged with the transcription files (which in some cases involved separate transcripts for separate speakers) and alignment was not always straightforward. Speaker labels retrieved through OCR (optical character recognition) --- as was the case for many of the theater pieces that were collected in \texttt{pdf} format --- were sometimes hard to recognize and match with text.

In many corpora, due to different discourse annotation objectives, a single speaker turn may be divided into clause-level units. These units had to be merged into whole turns for CFDD. In addition, some corpora represent overlapping speech by including embedded turns, such as, ``L: J'ai- Enfin j'ai bien aimé mais mais pas euh  \textbf{<E: Hum hum>} extrêmement quoi,'' in which E's utterance of ``hum hum'' overlaps with L's longer utterance (which translates roughly as ``I, I mean, I liked it pretty well but not extremely well''). It can also happen that a corpus exhibits annotation differences from one file to the next requiring individual ad-hoc interventions for individual files. This was the case in PFC, for example, although the special cases were fortunately easy to detect.

Different corpora also adopt different transcription and annotation conventions. Some might use parentheses, while others might use square brackets, angle brackets or special symbols to indicate non-vocal sounds or incomprehensible expressions (e.g., ``(dog barking)'', ``[clears throat]'', ``@'' for laughter). Anonymization is also indicated in heterogeneous ways, including ``[PII]'' (personally identifying information), ``NNAAMMEE'', ``BUZZ'', ``NPersonne`` or ``NLieu'' (where ``lieu'' means ``place'' in French), among others. Some transcripts contain punctuation and casing, while others do not. And finally,  corpora  differed with regard to the encoding of accented characters, the use of non-breaking spaces and the style of punctuation marks (e.g., " vs.\ ''). 

Some corpora include text that is useful for structuring dialogues, or transcripts of them, but that we did not wish to keep in the transcripts of CFDD. In the Assemblée Nationale discussions, for example, it is exceptionally common for a change in speaker to be explicitly marked by an utterance of ``La parole est à [name]'' (e.g., ``[name] has the floor''), where ``[name]'' is the name of the new speaker. We systematically removed speaker turns of this form when there was no other content in the turn. Likewise, the stage plays contain information about acts and scenes that was not kept in the final transcripts.

\subsection{Data normalization}

To normalize the datasets in CFDD, we opted for a simple text format in which each line of text corresponds to a new speaker turn, conversations are separated by a single blank line and special tags are consistently represented with square brackets.

\textbf{Speaker turns.} Each new line of text in Claire French corresponds to a single speech turn such that there are never two consecutive turns by the same speaker. Speaker labels are provided at the beginning of each line and are always formatted with square brackets and colons before the closing bracket.
When the identity of the speaker is unknown, we use "{\tt [speaker000:]}", "{\tt [speaker001:]}", "{\tt [speaker002:]}" and so on. Otherwise, the speaker can be designated by first name (e.g.,\,"{\tt [John:]}", "{\tt [Jean-Pierre:]}"), by first name and last name (e.g., "{\tt [Emmanuel~Macron:]}") or function (e.g.,\,"{\tt [Mr.\,le~President:]}", "{\tt [Interviewer:]}").

\textbf{Conversations.} The end of a conversation is always indicated by a single blank line. In most cases, the end of a conversation in the training data corresponds to the actual end of the transcript in the original raw data. For certain datasets, however, we were able to take advantage of their structure to divide them into smaller subsections which are then treated as ``conversations'' in the training data. To form conversations from the plays in Théâtre Classique and Théâtre Gratuit, for example, we arbitrarily chose to divide them into acts and scenes.
The political debates in FREDSum are divided by topic such as immigration, retirement, school, and so on.

\textbf{Special tags.}
We  standardized all annotations that indicate anonymization or signify non-vocal sounds using square brackets.
In the end, the normalized transcriptions do not include any parentheses, angle brackets or special annotation symbols.
We have also normalized some of the annotations that occur frequently in corpora, including:
\begin{itemize}
\item "{\tt [PII]}" (Personally Identifiable Information) for anonymized content. These are often names of people, and sometimes of places, dates, organizations, etc. This is the only bracketed annotation that corresponds to something that has been verbalized (but transcribed differently).
\item "{\tt [NOISE]}" for distinct ambient noises.
\item "{\tt [LAUGHTER]}" for laughter.
\end{itemize}

We have also standardized character encoding,
with NFC unicode standardization,
conversion of non-breaking spaces to single spaces,
conversion of ellipsis to three dots,
standardization of  apostrophes ({U+0027}) and quotation marks ({U+0022}),
etc.
As all the texts are in French, we have also forced French typographical rules to be followed, notably always a space before double punctuation marks ({\tt :}, {\tt ;}, {\tt !}, {\tt ?}).
These standardizations certainly remove diversity from the data,
but note that diversity can easily be recovered and generated through data augmentation, which gives greater control over what can be supported by models trained on our dataset.

\subsection{Train/test split}

Most of the datasets are split in two,
a train subset and a test subset. When we could find them, we followed the splits proposed by the original corpora. This is the case for 
Rhapsodie, Paris Stories, SUMM-RE and FREDSum. The goal here was to facilitate evaluation on downstream tasks and comparisons of different models. Someone wanting to test the impact of starting with a dialogue-pretrained model on the downstream task of summarization, for example, would need to use the same test set for SUMM-RE and FREDSum (both summarization corpora) when comparing a dialogue-pretraining approach to other models. 
In the event that we were unable to find available splits, we arbitrarily chose a few conversations from most of the other datasets in such a way that each category described in Section \ref{sec:data-description} was well-represented. Further information on how the datasets are split can be found on Hugging Face.\footnote{\href{https://huggingface.co/datasets/OpenLLM-France/Claire-Dialogue-French-0.1/blob/main/FR/metadata_split_testset_list.json}{\scriptsize https://huggingface.co/datasets/OpenLLM-France/Claire-Dialogue-French-0.1/blob/main/FR/metadata{\textunderscore}split{\textunderscore}testset{\textunderscore}list.json}}

\section{Looking forward}

CFDD is specifically designed to be used in the training of LLMs with an eye to downstream NLP tasks requiring the comprehension of French dialogue. This includes topics close to the research interests of LINAGORA and its associated LinTO project and open-source platform, such as the automatic generation of meeting minutes or query-based knowledge extraction from transcripts. But it should also have an interest for tasks requiring conversation understanding and generation more broadly.

It is also offered as a contribution towards building larger collections of open access, multilingual datasets that are ready to be used for LLM training. 
LINAGORA and our partners in OpenLLM France plan to continue pushing in this direction and releasing datasets in other genres and other languages. 
To this end, we invite input from the community regarding datasets that we may have missed but could be added to the CFDD\footnote{An issue, a pull request, or a discussion can be opened on \href{https://github.com/OpenLLM-France/Claire-datasets}{https://github.com/OpenLLM-France/Claire-datasets}}
and invite interested researchers to collaborate with us on future projects through OpenLLM France.\footnote{\href{https://huggingface.co/OpenLLM-France}{https://huggingface.co/OpenLLM-France}}\footnote{\href{https://github.com/OpenLLM-France}{https://github.com/OpenLLM-France}}


\end{document}